# Segmentation and Nodal Points in Narrative: Study of Multiple Variations of a Ballad

Fionn Murtagh (1) and Adam Ganz (2)
(1) Department of Computer Science
(2) Department of Media Arts
Royal Holloway, University of London, Egham, TW20 0EX, UK
fmurtagh@acm.org

**Abstract**

The Lady Maisry ballads afford us a framework within which to segment a storyline into its major components. Segments and as a consequence nodal points are discussed for nine different variants of the Lady Maisry story of a (young) woman being burnt to death by her family, on account of her becoming pregnant by a foreign personage. We motivate the importance of nodal points in textual and literary analysis. We show too how the openings of the nine variants can be analyzed comparatively, and also the conclusions of the ballads.

## 1 Introduction: Segmentation of Narrative Flow

In conscious thinking the conscious subject chops up both the what-is-thought and its expression into (let us call them to begin with) “chunks”. The linguistics work of Chafe (1994, 2002) has as its premise that “the flow of consciousness affects the flow and shape of language” and “it is the way we organize our thoughts that determines what we say”.

If not artificial or contrived, this flow takes place in time: “thoughts flow through time as well, and language is first and foremost a way of organizing and communicating this flow of thoughts. It is futile to limit our attention to isolated sentences. The shape a sentence takes can never be appreciated without recognizing it as a small, transient slice extracted from the flow of language and thought, when it has not simply been invented to prove some point.” (Chafe, 2002.)

Thoughts can be expressed in sentences but the essential point is that demarcating of the thoughts into “slices” or “chunks” is carried out: “At a higher level of organization, thoughts and language are formatted as a succession of topics. Each topic amounts to a partially activated cluster of knowledge within which speakers navigate with more limited, fully activated foci of consciousness. Speakers organize what they say in terms of basic-level topics, within each of which there can be a hierarchy of subtopics. At some level of this hierarchy we may encounter the kind of unit traditionally called a sentence, though I have found that sentence boundaries tend to be decided on the run as people talk, a fact which suggests that sentences may not reflect stored cognitive units of the same nature as topics and foci of consciousness. In any case the topic organization of thoughts and language needs to be studied as a fundamental determinant of discourse structure.” (Chafe, 2002).

Chafe (1979), in analyzing verbalized memory, used a 7-minute 16 mm colour movie, with sound but no language, and collected narrative reminiscences of it from human subjects, 60 of whom were English-speaking and at least 20 spoke/wrote one of nine other languages. Chafe considered the following units.

1. *Memory* expressed by a *story* (memory takes the form of an “island”; it is “highly selective”; it is a “disjointed chunk”; but it is not a book, nor a chapter, nor a continuous record, nor a stream).

2. *Episode*, expressed by a *paragraph*.

3. *Thought*, expressed by a *sentence*.

4. A *focus*, expressed by a *phrase* (often these phrases are linguistic "clauses"). Foci are "in a sense, the basic units of memory in that they represent the amount of information to which a person can devote his central attention at any one time".

The "flow of thought and the flow of language" are treated at once, the latter proxying the former, and analyzed in their linear and/or hierarchical structure by Chafe (1979), Hinds (1979), Longacre (1979) among others. For more general text, we can consider segmentation. Examples of text segmentation to open up the analysis of style and structure include Skorochod'ko (1972), Grosz and Sidner (1986), Hearst (1994), Bestgen (1998), Choi (2000) and Grosz (2002).

We have here various ways of defining or determining units of thinking and of narrative. In this work, we pursue a bottom-up approach, taking the textual data as our starting point. It is often surprising how far we can go in reducing any text to the words, alone, that are used in it. Consider for example a key term used in a politician's speech, or common word patterns used for author attribution.

General patterns of word usage provide for genre analysis. So-called tool or function words (for example: to, and, the, etc.) also have their role to play, in providing useful pointers to style and, as a knock-on effect, allowing for emotional positioning in the storyline of the text. See Murtagh et al. (2009) for more discussion on this.

In this work we use a so-called bag of words approach, taking all words, as individual words, into account. In order to avoid a need for harmonization, we ignore punctuation entirely, and also force all upper case letters to be lower case. Finally we impose a lower acceptable limit of two letters to define a word. The latter implies that the indefinite article, the personal pronoun, and words that comprise one letter through having punctuation removed, are all ignored in our analysis. On the basis of many studies over the years, and also basing our work on a tradition in text analysis that has been used over the past half century – see Lebart et al. (1998), Murtagh (2005), Le Roux and Rouanet (2010) – we adopt the approach used here.

Our aims are to study segmentation of narrative, from the text, and simultaneous with segmentation to specify nodal points in the narrative. The questions we pose are:

1. How well can we devise an algorithm to automatically segment a narrative, baselined against external judgement on appropriate segment boundaries?
2. On the same basis of expert knowledge, do nodal points fall on segment boundaries?
3. Assuming that our findings support the two previous issues, then if we are given variants of a narrative, do the segmentations and nodal points found for each variant match well with the respective variants of the narrative?

We use the Lady Maisry ballads since they provide a set of related variants of much the same story. The data is available at http://sacred-texts.com/neu/eng/child/ch065.htm The ballad from the Francis J. Child collection (Nelson-Burns, 1999) deals with the burning to death by her family of a Scottish woman who has become pregnant by an English lord.

For research questions 1 and 2 above, we used ballad 65B, discussed from a literary and cultural point of view in Ganz (2010). For research question 3 we use in addition ballads 65A, 65C, 65D, 65E, 65F, 65G, 65H and 65K. These versions are all (reasonably) complete. In 65K, two stanzas have repeated sequence numbers and we simply renumbered sequentially.

## 1.1 Nodal Points in Screenplay Analysis

Nodal points are important for script writing, in that they present place in a narrative where different possibilities or threads of action intersect. Field (1994) in his hugely influential text *Screenplay* identifies key points of action and approximately where they should occur in a screenplay. This approach is often criticized for being over-prescriptive or formulaic.  Yet there is an increasing tendency that scripts which are made into films are written to the formula, and those films which are deemed to be of acceptable standard are those that adhere to the main tenets of Field's theory.

Our work makes it possible for a given text to reveal its own nodal points, which may then be compared with other variants of that text or with similar examples of the genre to improve an analysis of script structure, and to see where those points of intersection can be said to naturally occur. We can identify the places in which one part of a story connects with or responds to other parts and by mapping those nodal points we are closer to developing a sense of narrative structure as a specific pattern composed of the totality of these intersections which may be distinctive for a particular author, genre, or period.

# 2 Analysis of Lady Maisry Ballad 65B

As we note in Ganz (2010), "Often ballads – and films – rather than working around a clear narrative, work more around a cluster of nodal point story points. In this [65B: Lady Maisry] ballad we can see three of these clusters –

Janet's discussion with her family, the servant's ride to London and the Lord's return, and the bloody coda."

In this study, we see if or how we can find such nodal points directly from the text.

We use a "bag of words" approach to characterize text segments. Each such text segment or unit is given by a stanza. In the 27 stanzas in ballad 65B there are 675 words. We used 210 unique words of length two or more characters, with punctuation ignored, and all upper case converted into lower case. The most frequent words, with overall frequency of occurrence, are: the 24, to 24, he 22, and 21, whore 16, that 14, her 13, in 11, me 11, for 10, your 10, be 9, on 9, brother 8, is 8, my 8, you 8, janet 7, ll 7, etc. Note (last entry listed) how the deletion of the apostrophy in "I'll" means that "ll" is left as a word in our sense.

The ballad form is an ideal form of text to use. Repetition, which is refrain-like or otherwise provides "binding" or continuity in the story, can be neatly counterposed, as we will show, to non-continuity or change when a (somewhat) new word set is used in a following stanza.

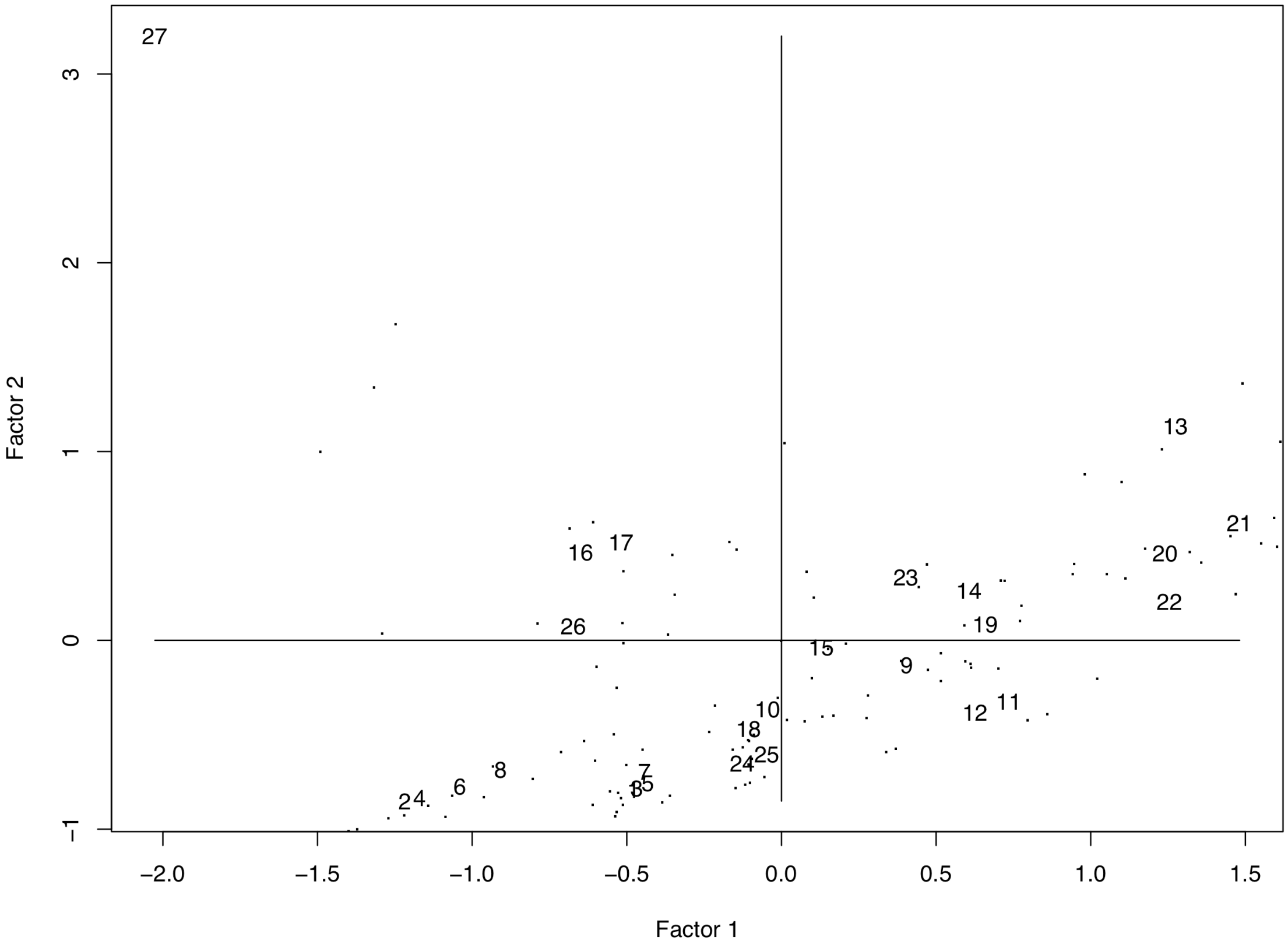

Figure 1: Correspondence Analysis principal factor plane of 27 stanzas crossed by 210 words contained in them. The dots are the locations of the words. The 27 stanzas are located at, respectively, 1, 2, ... , 27. The importance of these factors are given by the percentage inertia explained: respectively, 5.7% and 5.4% for factor 1 and factor 2.

Fig. 1 shows a best planar display of ballad 65B, where we have taken the set of stanzas in a 210-dimensional (word) space, and the set of words in a 27-dimensional (stanza) space, and found the best fitting planar (2-dimensional) projection. See Murtagh (2005) for background description on the

Correspondence Analysis methodology.

The inherent dimensionality is 26 (this is the minimum of 210 and 27, less 1 because of a linear dependence that is part and parcel of the embedding algorithm). In this new, factor space, we have the stanzas and also the words in a common 26-dimenensional space.

Should we in preference use the 2-dimensional space seen in Fig. 1, the full – hence no loss of information – space that is 26-dimensional, or some "cleaned" version in between? We looked at the succession of percentages of inertia associated with factors. We found interest in an intermediate number of dimensions, 15, because immediately thereafter there is a somewhat greater relative drop in importance as quantified by the percentage inertia explained. We then proceeded to checking the nodal point, or relatively large change, situation therefore for dimensionalities of 2, 15 and 26. For this we use hierarchical clustering, yielding the dendrogram or tree diagrams shown in the next and subsequent sections of this article. Happily we find very little difference in the outcome, implying that the reduced dimensionality mapping is not overly sensitive to how reduced the embedding happens to be.

In our work now we will use the 2-dimensional embedding – as in Fig. 1 – because it furnishes visually more differentiated branchings in the hierarchical clustering built from it. We can say informally that we are using Correspondence Analysis for data cleaning or filtering when we retain in this way just two underlying dimensions, on which to pursue our aims of finding segments and nodal points.

## 3 Change Detected Using Sequence-Constrained Hierarchical Clustering

We use a hierarchical clustering that is carried out on the sequence of stanzas. The agglomerative clustering criterion is the complete link one. It is proved in Murtagh (1985) that this criterion, when subject to the constraint of only allowing adjacent clusters to be agglomerated, is guaranteed to be well-behaved in the sense of guaranteeing no inversions. It was used to great effect in, e.g., Murtagh et al. (2009, 2010) on film and television script studies.

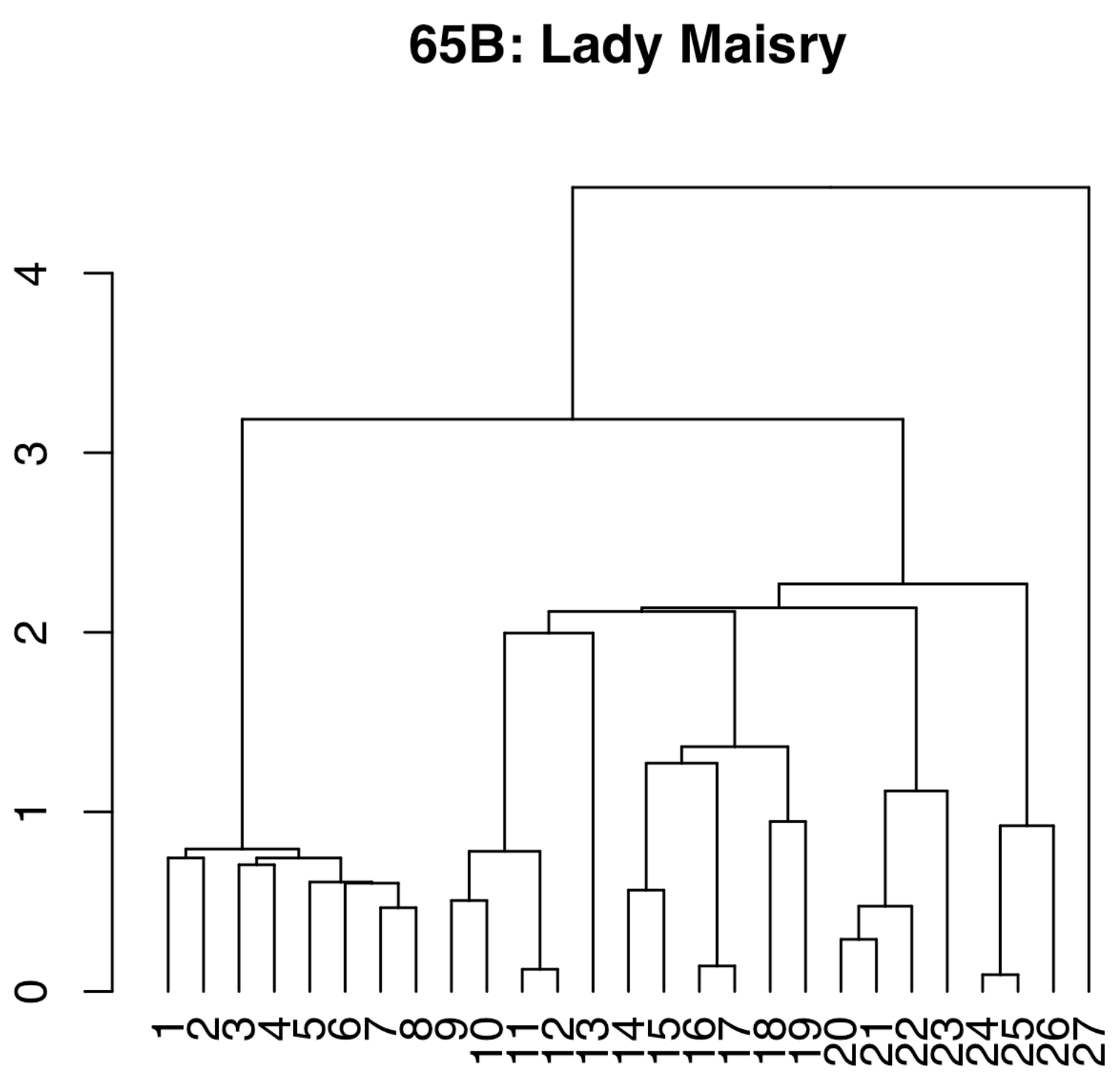

Figure 2: Hierarchical clustering, constrained to be on the succession of 27 stanzas. Here the stanzas are in a 2-dimensional projection, as shown in Fig. 1.

Looking at Fig. 2 we note the following.

- Stanzas 1–8: Janet's discussion with her family.
- Stanzas 9–12: old woman, fire, question about servant, locating servant.
- Stanza 13: servant on the way to the English lord.
- Stanzas 14–19: servant with the lord, and explanation given.
- Stanzas 20–23: lord on the way to Lady Maisry, being burned.
- Stanzas 25–26: at the scene of the burning.
- The final stanza, 27, is found to be very different from the preceding ones. It relates to future revenge.

We can read off segments on different scales, for example the ones listed in the foregoing list. Moving up the hierarchy gives us as a lower resolution alternative:

- 1–8 (scene setting, allegations and refutation),
- 9–23 (organizing and seeking rescue from the English lord),
- 24–26 (at the scene of the burning), and
- 27 (anomalous stanza, future-oriented).

The turning points, providing also for nodal points, are at the transitions between these segments.

## 4 All Segmentation and Nodal Points in Other Lady Maisry Ballad Variants

In ballad 65A, see Fig. 3, stanza 12 follows the introduction of Maisry rejecting local suitors, being pregnant from the English lord, and a brother who says that she will die for that. In stanza 12, her plight and pending murder becomes known to her. We see in Fig. 3 that the ending is less different from previous stanzas, compared to ballad 65B discussed in the previous section above. From 65A, one low resolution segmentation that we can read off the dendrogram is up to stanza 11, and then all the action is from stanza 13 onwards. Within the latter main action part, we can distinguish sub-plots, starting with the part up to stanza 20, and then from stanza 21 onwards. It is from stanza 21 that help is being sought from the English lord, to save Maisry.

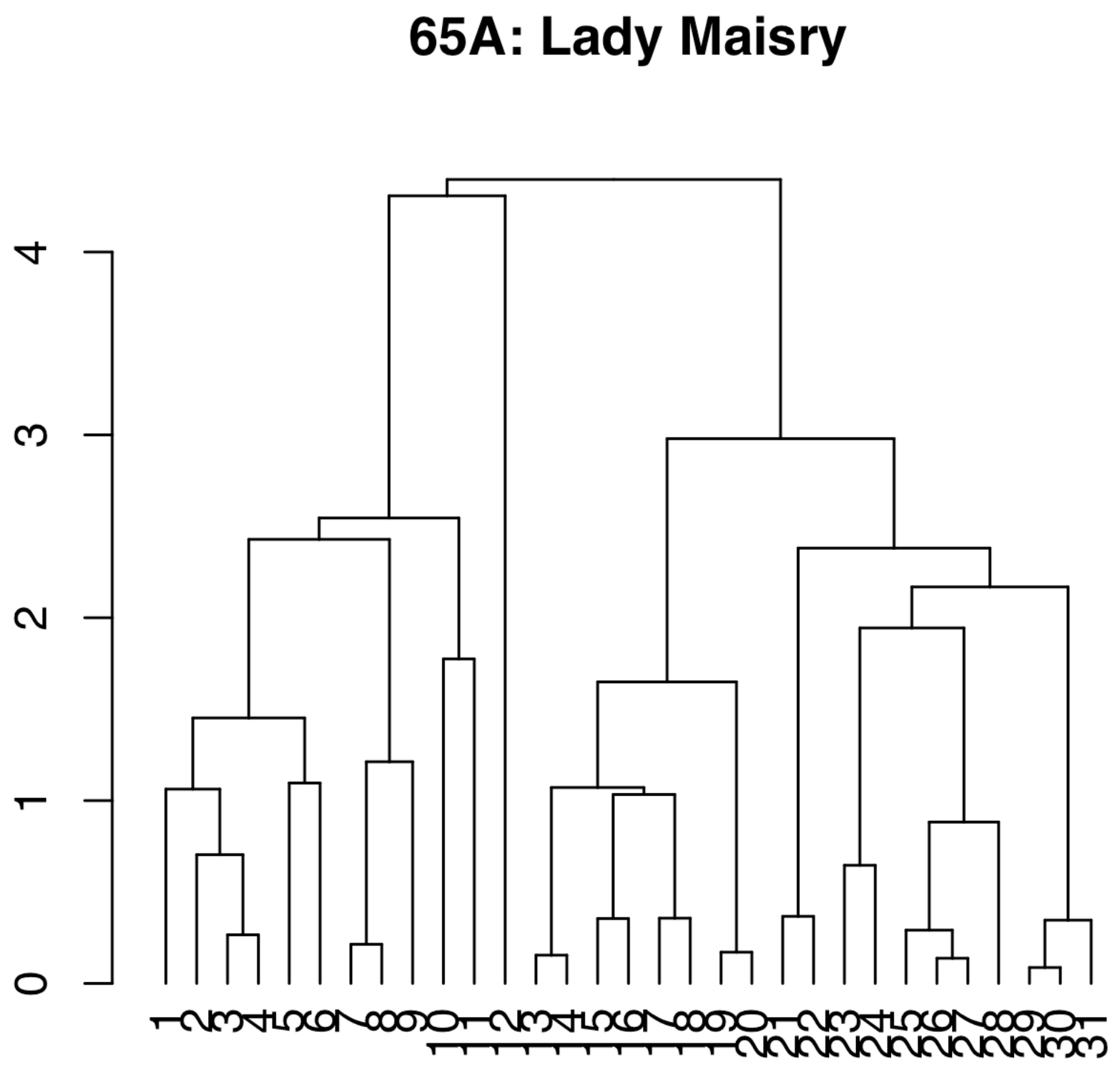

Figure 3: Hierarchical clustering, constrained to be on the succession of stanzas, based on a 2-dimensional projection from Correspondence Analysis.

In ballad 65C, see Fig. 4, seeking aid is underway by stanza 7. Stanzas 21 and 22 both point to the future, to revenge and to making the messenger boy an heir of the English lord. Even if not a partition defined from one level (horizontal cut) of the dendrogram, a fairly low resolution level segmentation is up to stanza 6, from stanza 7 to stanza 20, and then the final cluster comprising stanzas 21 and 22.

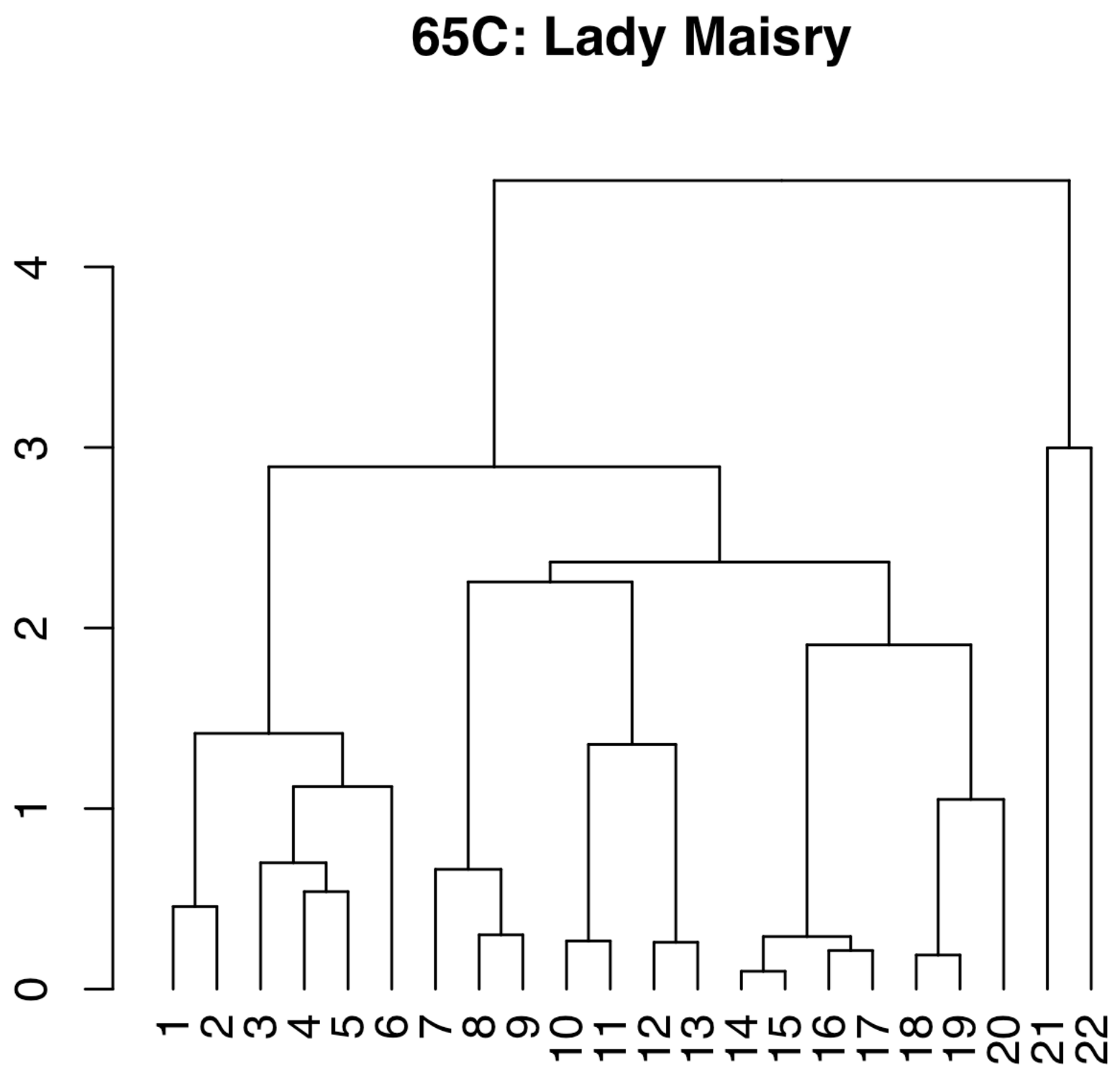

Figure 4: Hierarchical clustering, constrained to be on the succession of stanzas, based on a 2-dimensional projection from Correspondence Analysis.

In ballad 65D, see Fig. 5, there is a fairly pedestrian statement of the facts in stanzas 1–5. Stanza 15 onwards is the English lord on his way. Stanza 18 seems overly atypical in the dendrogram because it describes colourfully the haste of the lord on his way. By stanza 19 he is with the dying Maisry. While thereafter revenge is planned, this is over all remaining stanzas, especially the last three.

**65D: Lady Maisry**

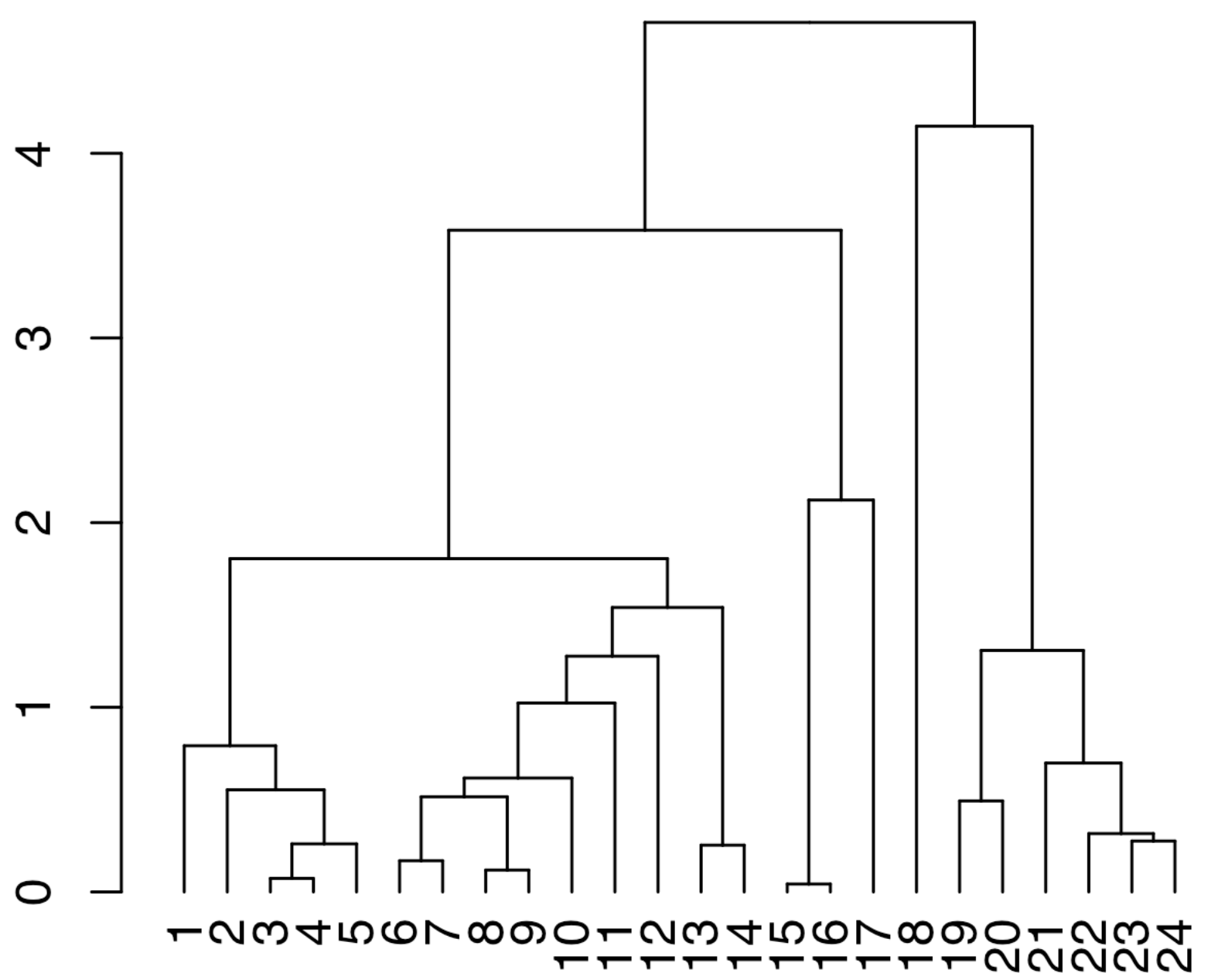

Figure 5: Hierarchical clustering, constrained to be on the succession of stanzas, based on a 2-dimensional projection from Correspondence Analysis.

In ballad 65E, see Fig. 6, stanza 15 is where the dying Maisry hears a horn presaging the arrival of the English lord. Stanzas 16 and 17 indicate Maisry's hope, although in stanza 18 the lord arrives but she is dead.

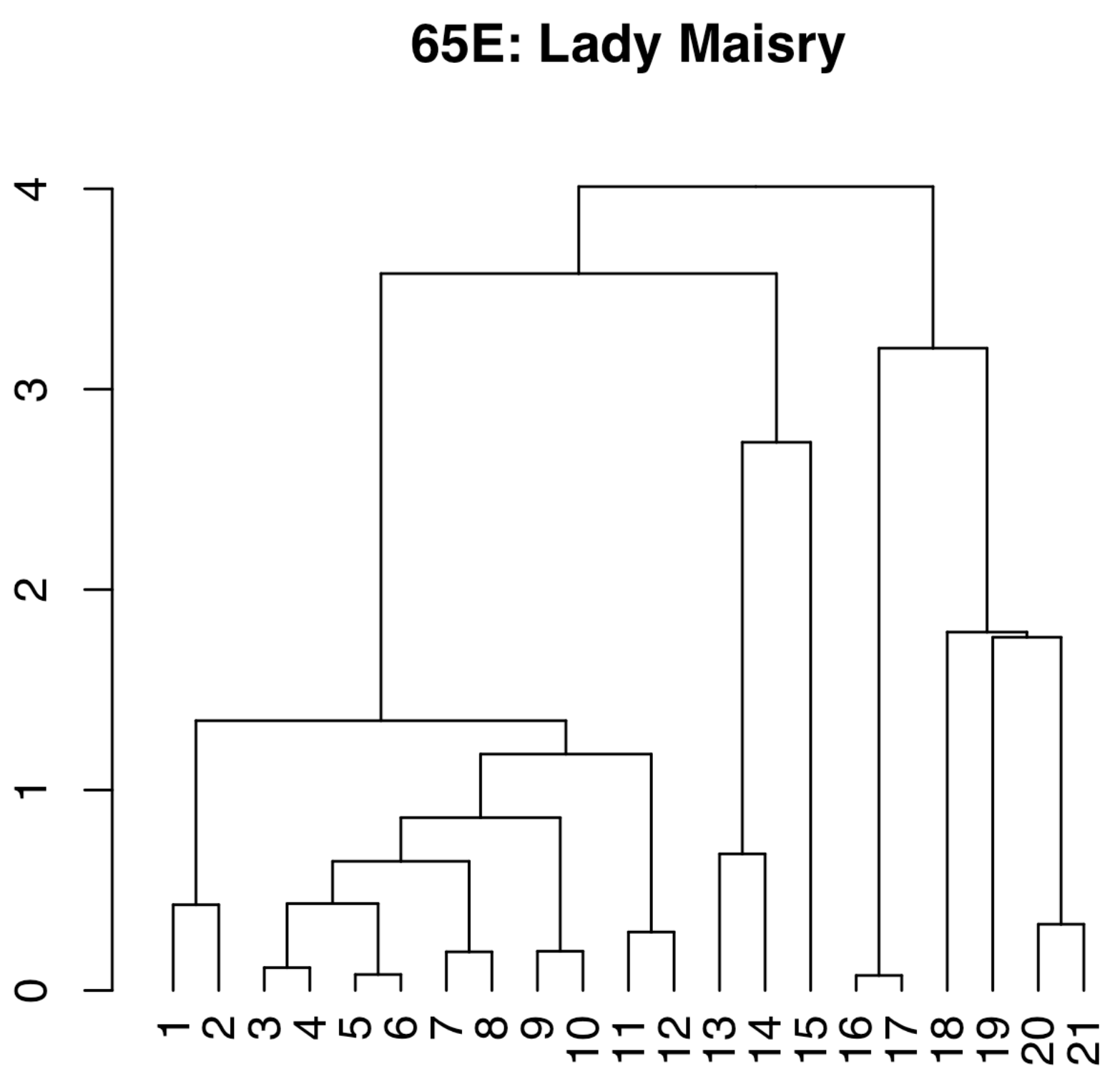

Figure 6: Hierarchical clustering, constrained to be on the succession of stanzas, based on a 2-dimensional projection from Correspondence Analysis.

Ballad 65F, see Fig. 7, has our sequencing, and we have omitted the original (and nearly empty) stanza 11. In our sequencing, as used in the figure, from stanza 13 onwards the messenger boy is on his way. Stanza 19 in our sequencing is where the dying Maisry (referred to as Marjory in this variant) hears the horn of the English lord.

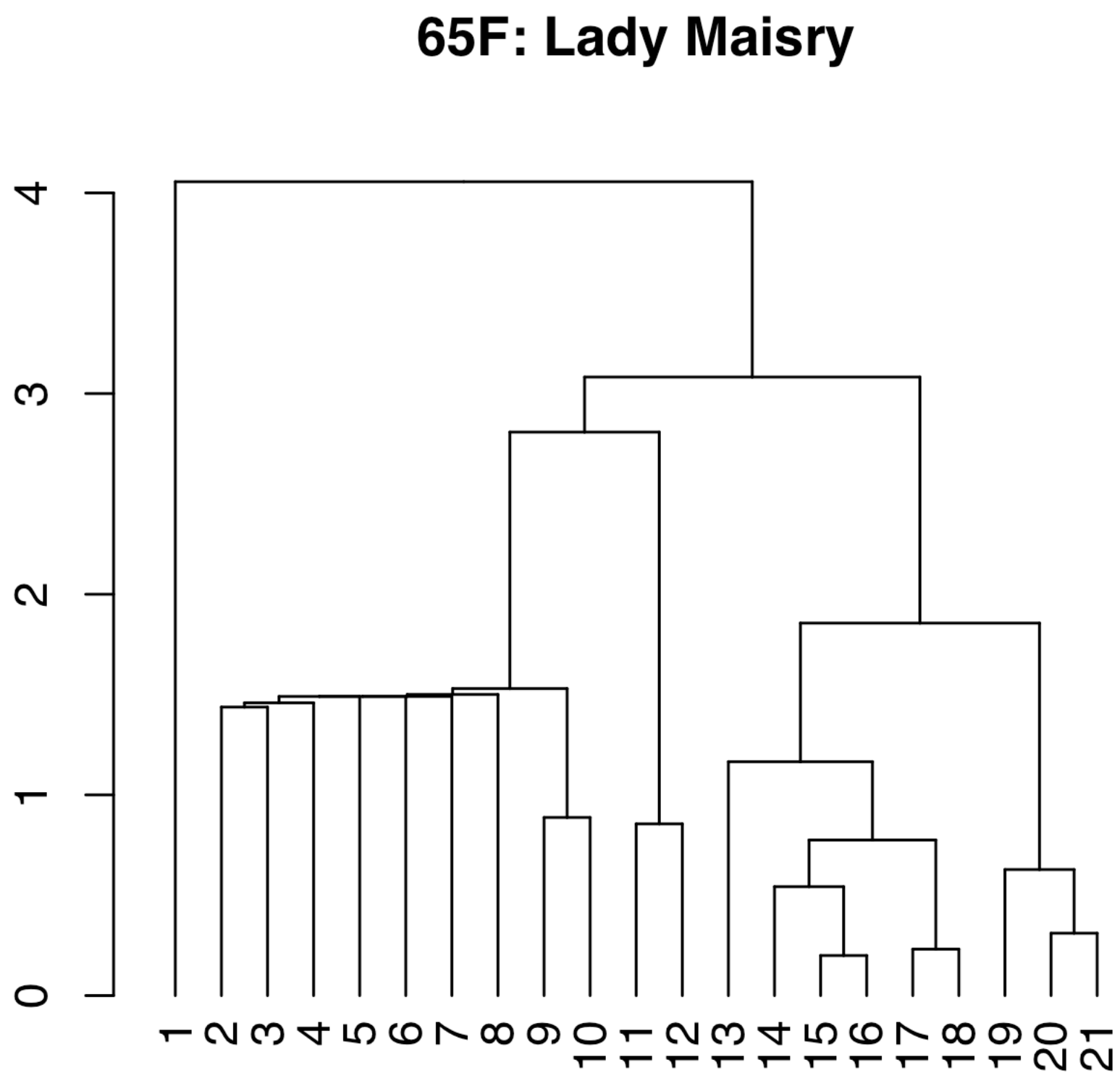

Figure 7:  Hierarchical clustering, constrained to be on the succession of stanzas, based on a 2-dimensional projection from Correspondence Analysis.

In ballad 65G, see Fig. 8, the foot-page – messenger boy – is underway in stanzas 4–7. By stanzas 10–11, the English lord is underway. In stanza 15, the last, Maisry dies, in spite of hopes raised in the previous stanzas.

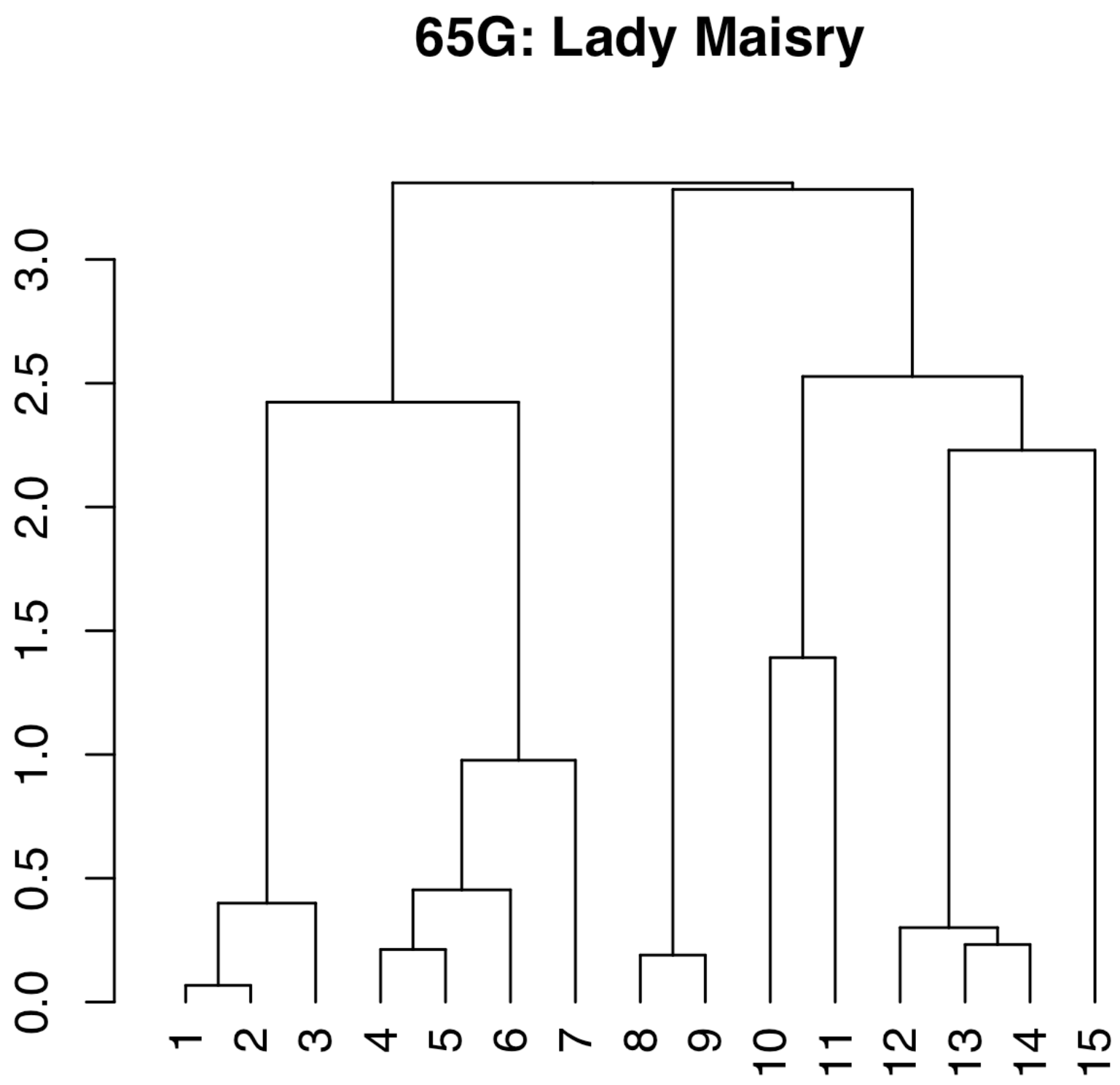

Figure 8: Hierarchical clustering, constrained to be on the succession of stanzas, based on a 2-dimensional projection from Correspondence Analysis.

In ballad 65H, see Fig. 9, it is in stanza 14 that Maisry's brother says he will kill her, and in subsequent stanzas proceeds to do so. In stanza 20, in this case the English prince James, suspects that something is up and seeks a messenger boy to get information. In stanza 27, news is brought back about the preparations for the killing of Maisry. Stanza 29 onwards is Maisry knowing before she dies that James is on the way, his arrival and his planning of revenge.

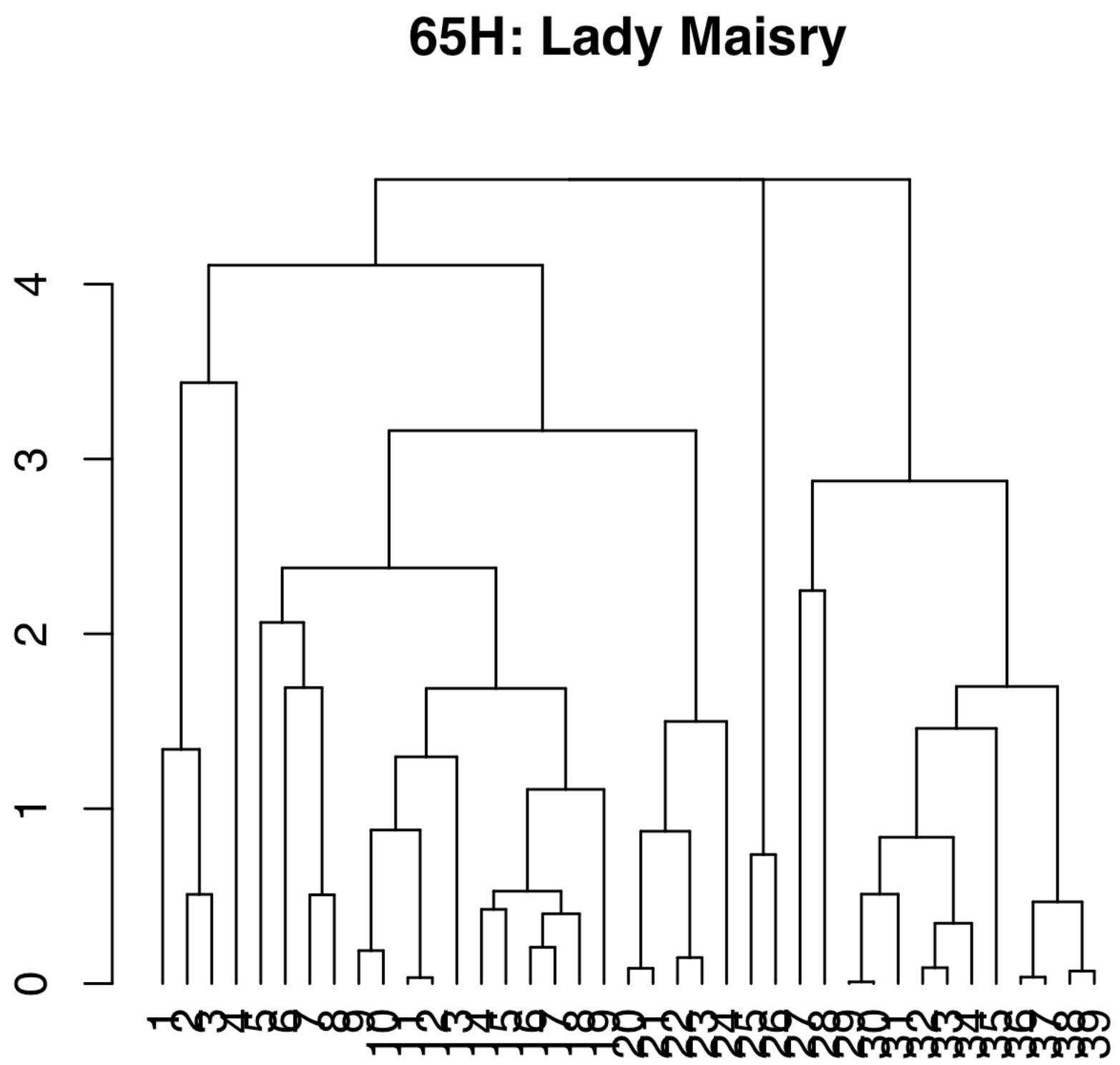

Figure 9: Hierarchical clustering, constrained to be on the succession of stanzas, based on a 2-dimensional projection from Correspondence Analysis.

In ballad 65K, see Fig. 10, stanza 15 (using our sequencing in succession of the 19 stanzas) sees a non-contrite Maisry being burned alive by her father and her mother. Before that is the buildup to the killing scene. After stanza 15 is the arrival of the lord, her lover, and his promises of revenge. Meanwhile stanzas 1 and 2 were fairly gentle background introductions.

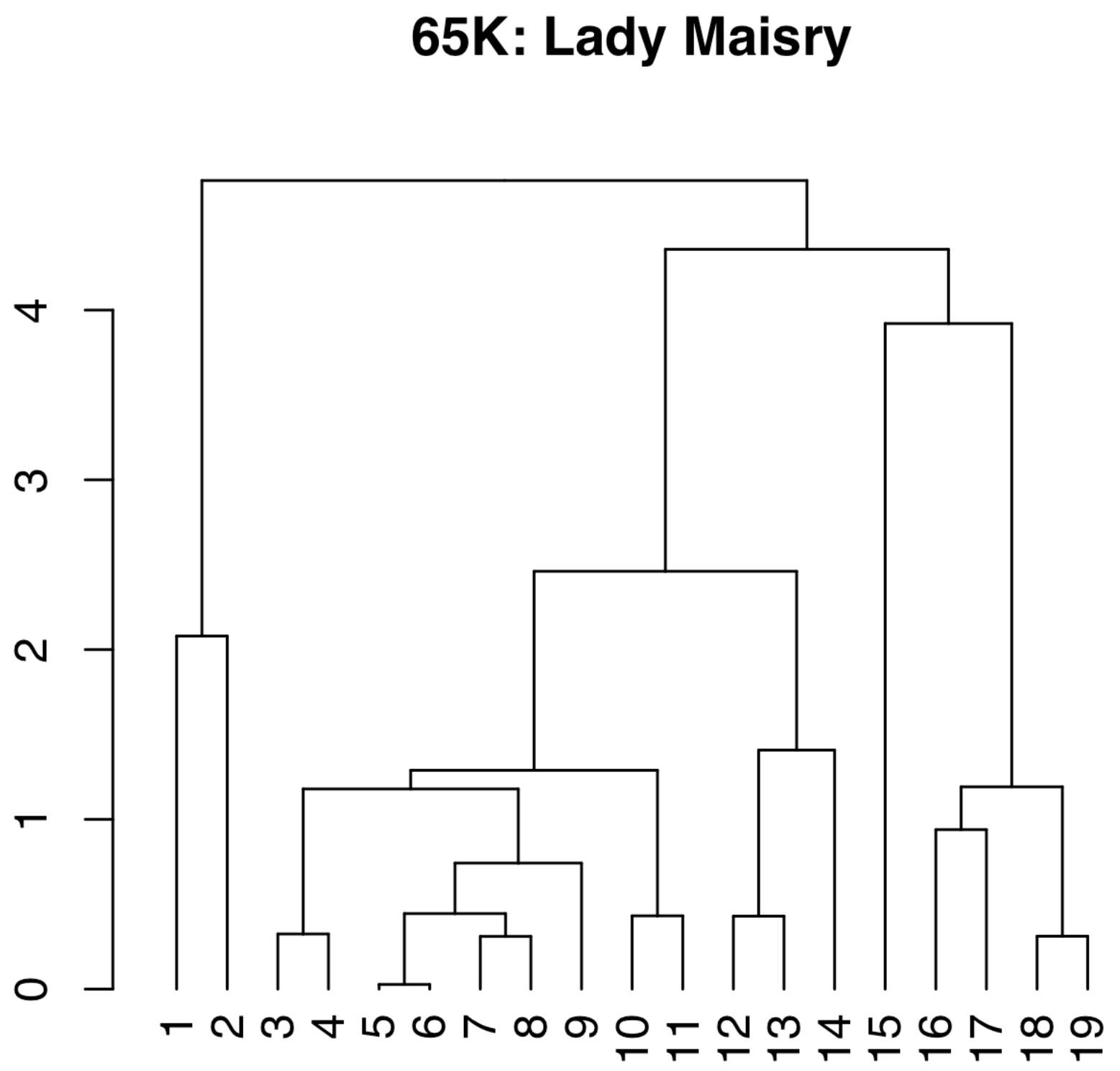

Figure 10: Hierarchical clustering, constrained to be on the succession of stanzas, based on a 2-dimensional projection from Correspondence Analysis.

All variants of the Lady Maisry theme, as displayed in the dendrograms, show how we can define segments, that have meaningful nodal points. We can do this at different resolution levels, based on the embedding of clusters that are portrayed in the dendrogram. As shown, we produce in this way very credible points of change, and of continuity, in the storyline. We have referenced these segmentation and nodal point conclusions to the original ballad texts.

# 5 Analysis of All 9 Selected Variants

In Table 1 we summarize the properties of the Lady Maisry ballad variants that we used. The analyses carried out thus far are based on the set of stanzas and the set of unique words.

| Variant | No. stanzas | Total words | Unique words |
|---|---|---|---|
| 65A | 31 | 809 | 288 |
| 65B | 27 | 675 | 210 |
| 65C | 22 | 559 | 217 |
| 65D | 24 | 596 | 232 |
| 65E | 21 | 577 | 210 |
| 65F | 21 | 530 | 197 |
| 65G | 15 | 400 | 166 |
| 65H | 39 | 993 | 312 |
| 65K | 19 | 471 | 196 |
| all | 219 | 5610 | 837 |

Table 1: Nine variants, A through K, of the Lady Maisry ballad were used. In F we omitted one (nearly empty) stanza and the number of stanzas here takes account of this. The total number of words are as measured by the Unix wc (word count) command, i.e. adjacent tokens. The numbers of unique words are based on: upper case converted to lower case; punctuation and numerical figures removed; and more than two character adjacent token groups.

In Fig. 11 we show the overall view based on the 9 variants of the ballad studied. These 9 variants contained 291 stanzas. The exceptional stanzas in the upper right are from 65F. In this variant, Maisry becomes Marjory, “mother” is “mither”, “become” is “becum”, “floor” becomes “flure”, and so on.

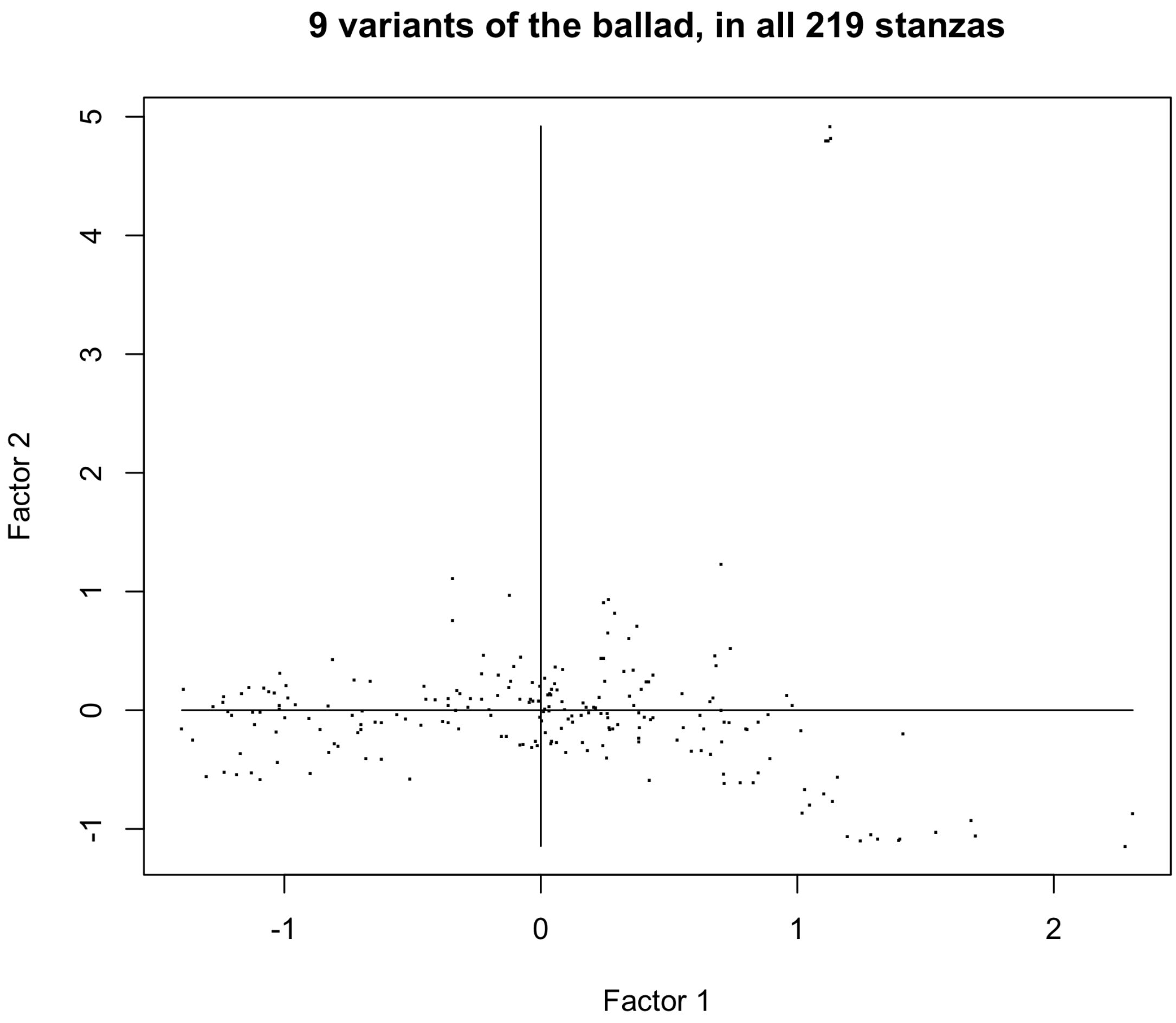

Figure 11: Correspondence Analysis principal plane projection, i.e. best two-dimensional view of the data, based on all data described in Table 1. Each dot in the figure represents a stanza, from the 219 stanzas overall that are analyzed here. The characterizing word set used 837 words (cf. Table 1).

In Fig. 12 we look at how the ballad variants start. A starts with Maisry’s disinterest for a Scottish partner. In B, Maisry’s sister suggests she is a whore. In C, her father does. In D, Maisry, here Margery, is away to Strawberry Castle, and similarly in E. In F, now Marjory, is “richt big wi bairn” (very big with child). G starts but in a non-shocking way with mother and father killing Maisry, in Maisry’s own words. H is relaxed: Maisry is sitting, sewing. K is like D and E. Fig. 12 shows that H is quite distinct in its opening, followed by F and G. In Correspondence Analysis, the origin (the location with coordinates (0,0)) is the average, and this is often enough the “semantic” or unremarkable average too.

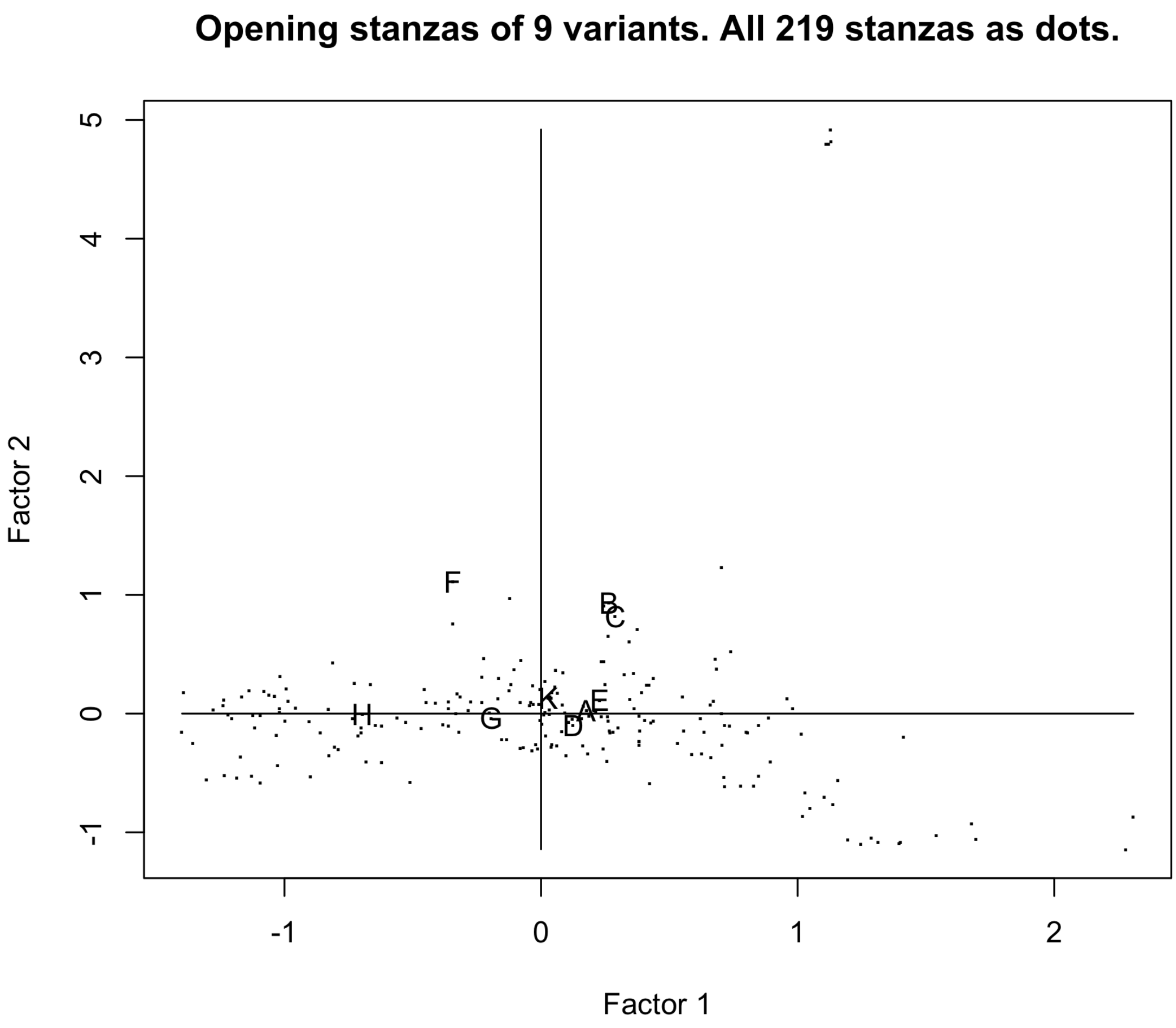

Figure 12: Opening stanzas of the 9 ballad variants are shown.

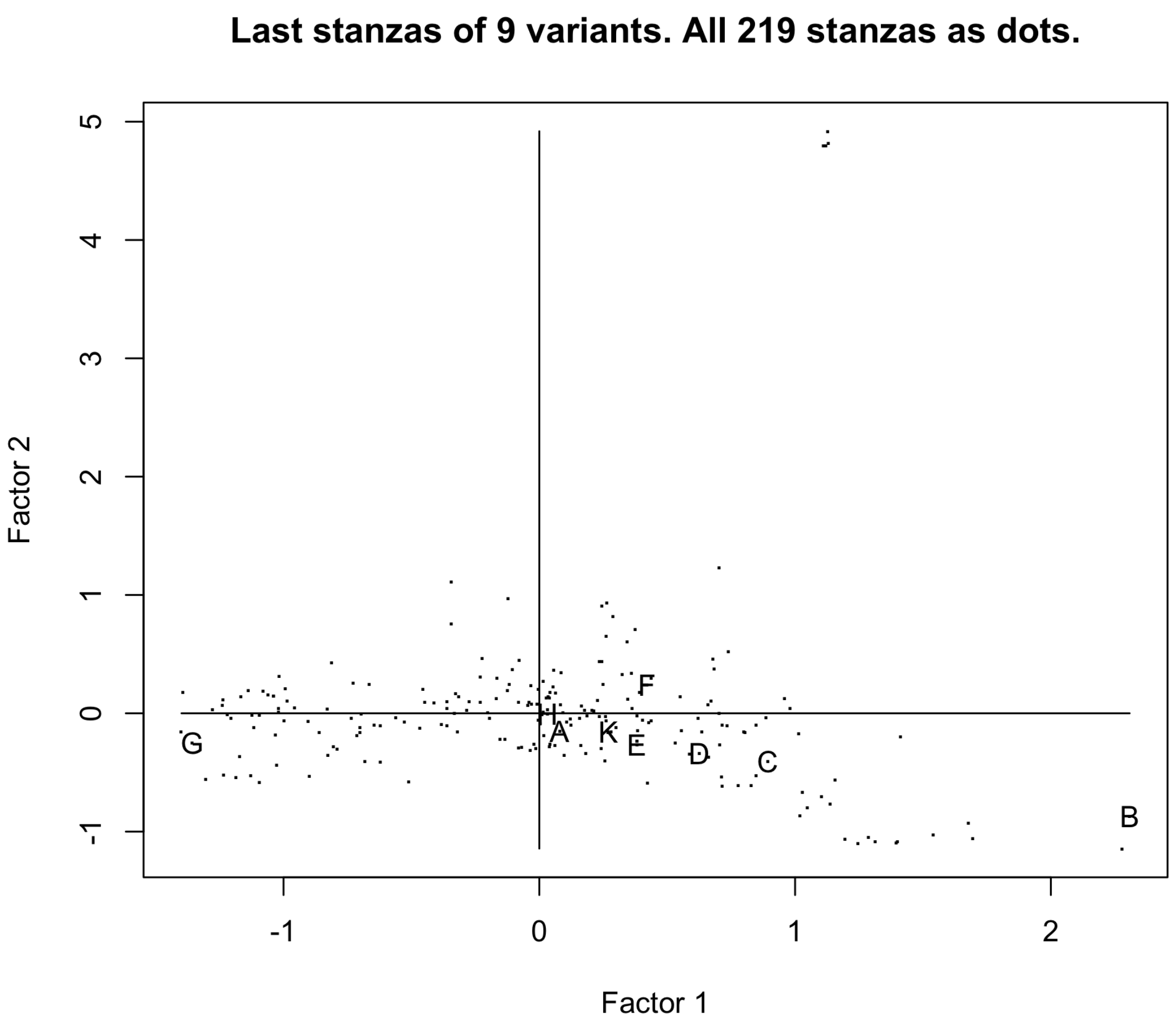

Figure 13: Final stanzas of the 9 ballad variants are shown.

In Fig. 13, we see that the final stanzas of B and G are quite different from the others. In fact the distinction between B and G in a way is fundamentally important for the overall set of 219 stanzas.

In A, the last stanza has the lord, Maisry's lover, threatening revenge on Maisry's kin, and his own suicide. In B, he threatens all round slaughter. In C, he will make the errand boy his heir. In D, all round slaugher is promised. In E, slaughter is promised and the errand boy to be remembered. F is like A. G indicates a vain attempt to save Maisry and suggests he dies in the attempt. In H,

Maisry is remembered and her lover is mad angry (“for he ran brain”). In K we are back to slaughter promised.

## 6 Conclusion

In Figs. 2-10 we can allow around six successive segments per ballad, taking into account unusual segments with one or a small number of stanzas. From these figures, we have a presentation of introductions to the story, of the scene-setting, of the attempt by Maisry to save herself from her murderous family and local society, of the attempt by her lover to come to her aid, and of the dénouement for her. We also have the various conclusions. So how much continuity there is in these parts of the story, or how much discontinuity, can be appreciated in these figures.

In Figs. 12 and 13 we look at all ballad variants to see which ones are somewhat more or less exceptional in terms of openings and closings. We could similarly take some other theme, such as Maisry’s seeking rescue, and average the relevant stanza in each ballad. This would also allow us to see what ballads are similar and which are different from the others and to create a taxonomy for the ballads based on this structural similarity or difference.

Overall we have a powerful means of comparative choice – of displaying a variant and relating it to its alternatives.

## References

Bestgen, Y. (1998). Segmentation markers as trace and signal of discourse structure, *Journal of Pragmatics*, 29, 753–763.

Chafe, W.L. (1979). The flow of thought and the flow of language, In *Syntax and Semantics: Discourse and Syntax*, ed. Talmy Givón, vol. 12, 159–181, Academic Press.

Chafe, W. (1994). *Discourse, Consciousness, and Time: The Flow and Displacement of Conscious Experience in Speaking and Writing*, Chicago: University of Chicago Press.

Chafe, W. (2002). Searching for meaning in language: a memoir, *Historiographia Linguistica*, 29, 245–261.

Choi, F.Y.Y. (2000). Advances in domain independent linear text segmentation, Proceedings of the First Conference of the North American Chapter of the Association for Computational Linguistics (Seattle, WA), ACM International Conference Proceedings Series Vol. 4, pp. 26–33.

Field, S. (1994), *Screenplay*. 3rd edn., New York: Dell.

Ganz, A. (2010). Time, space and movement: screenplay as oral narrative, *Journal of Screenwriting*, 1 (2), 225–237.

Grosz, B.J. (2002). Discourse structure, intentions, and intonation, in *The Languages of the Brain*, ed. A Galaburda, S Kosslyn and Y Christen, Harvard University Press, Cambridge, pp. 127–142.

Grosz, B.J. and Sidner, C.L. (1986). Attention, intentions, and the structure of discourse, *Computational Linguistics*, 12, 175–204.

Hearst, M. (1994). Multi-paragraph segmentation of expository text, Annual Meeting of the ACL, Proceedings of the 32nd Annual Meeting on Association for Computational Linguistics (Las Cruces, New Mexico), (Association for Computational Linguistics, Morristown, NJ) pp. 9–16.

Hinds, J. (1979). Organisational patterns in discourse, in *Syntax and Semantics, Discourse and Syntax*, ed. Talmy Givón, pp. 135–157, Academic Press.

Lebart, L., Salem, A. and Berry, L. (1998). *Exploring Textual Data*. Kluwer. (Lebart, L. and Salem, A., 1994, *Statistique Textuelle*, Dunod.)

Le Roux, B. and Rouanet, H. (2010), *Multiple Correspondence Analysis*, Sage.

Longacre, R.E. (1979). The paragraph as a grammatical unit, In *Syntax and Semantics: Discourse and Syntax*, ed. Talmy Givón, vol. 12, pp. 115–134, Academic Press.

Murtagh, F. (1985). *Multidimensional Clustering Algorithms*. Physica-Verlag.

Murtagh, F. (2005). *Correspondence Analysis and Data Coding with R and Java*. Chapman and Hall.

Murtagh, F., Ganz, A. and McKie, S. (2009). The structure of narrative: the case of film scripts. *Pattern Recognition*, 42, 302–312.

Murtagh, F., Ganz, A., McKie, S., Mothe, J. and Englmeier, K. (2010). Tag clouds for displaying semantics: the case of filmscripts, Information Visualization Journal, in press.

Nelson-Burns, L. (1999). Francis J. Child Ballads, Biography, Lyrics, Tunes and Historical Information, web site, 7 March 1999,

http://www.contemplator.com/child (accessed 2 April 2010).

Skorochod'ko, E.F. (1972). Adaptive method of automatic abstracting and indexing, Proc. of IFIP Congress 71, pp. 1179–1182.